# Certainty Modeling of a Decision Support System for Mobile Monitoring of Exercise-induced Respiratory Conditions

Chinazunwa Uwaoma
The University of the West Indies, Jamaica
chinazunwa.uwaoma@mymona.uwi.edu

Gunjan Mansingh
The University of the West Indies, Jamaica
gunjan.mansingh@uwimona.edu.jm

## Abstract

*Mobile health systems in recent times, have notably improved the healthcare sector by empowering patients to actively participate in their health, and by facilitating access to healthcare professionals. Effective operation of these mobile systems nonetheless, requires high level of intelligence and expertise implemented in the form of decision support systems (DSS). However, common challenges in the implementation include generalization and reliability, due to the dynamics and incompleteness of information presented to the inference models. In this paper, we advance the use of ad hoc mobile decision support system to monitor and detect triggers and early symptoms of respiratory distress provoked by strenuous physical exertion. The focus is on the application of certainty theory to model inexact reasoning by the mobile monitoring system. The aim is to develop a mobile tool to assist patients in managing their conditions, and to provide objective clinical data to aid physicians in the screening, diagnosis, and treatment of the respiratory ailments. We present the proposed model architecture and then describe an application scenario in a clinical setting. We also show implementation of an aspect of the system that enables patients in the self-management of their conditions.*

## 1. Introduction

There is no doubt that advanced technologies have helped reduce problems associated with managing long-term illnesses, and shortages of health professionals. Respiratory disorders such as exercise-induced asthma (EIA) or exercise-induced bronchospasm (EIB), exercise-induced rhinitis (EIR), exertional vocal cord dysfunction (VCD), as well as chronic obstructive pulmonary disease (COPD), are among the long-term health conditions whose management and treatment can benefit from real-time and continuous monitoring; given their social, emotional and economic impact on active and competitive individuals as well as the general populace [1, 2, 3, 4]. Several studies however, have reported that exacerbation of respiratory conditions due to strenuous activities are not well characterized during diagnosis [5]; the main reason being the high similarity and overlap in the triggers and symptoms particularly between EIB/EIA and VCD. Comorbidity cases in an individual also contributes to the under-diagnosis and under-treatment of these conditions [5, 6, 7].

Though there is no standard for the management and control of respiratory disorders, each ailment has an acceptable measure of wellness [8, 9]. Many researchers have focused on the use of advanced information and communications technologies to improve overall management and control of respiratory health conditions [10, 11]. Mobile health monitoring based on pervasive data capturing, presents patients the opportunity to personally monitor and manage their conditions as it enables real-time and continuous monitoring, and management of the conditions.

Mobile phones are not only capable of recording breath sounds [7, 11, 12], but also, able to perform analysis on the recorded signal. Wheezes, stridor, cough are frequent symptoms experienced by patients with exercise-induced respiratory conditions (EIRC); and doctors have identified these sounds as principal signals of respiratory distress [7]. However, lung or breath sound detection is not sufficient to provide absolute functional solution to pulmonary exacerbation. There is need to measure and analyze other vital signs including environmental effects as well as patient's level of activity, to provide accurate and reliable information in controlling the respiratory conditions [8,13].

Our study objective centers on developing a simple and portable respiratory monitoring system that can detect vital signs, perform signal analysis and context recognition; and also, send alerts and feedback to the user in real-time scenarios. The design architecture is based on machine learning for classification of respiratory sounds [7], and DSS for context recognition using expert system frameworks [14]. The inferencing mechanism of the embedded expert system is built on inexact reasoning- a concept of certainty theory that allows inferencing based on the level of belief in the presented evidences or in available contextual information. We deem this approach appropriate for the envisaged monitoring system given the overlap in the triggers and symptoms of exercise-induced respiratory conditions which of course, introduces some degree of uncertainty in their assessment. Uncertainty or approximate reasoning becomes necessary when the system is presented with situational evidences that require further probing.

Studies on modeling of uncertainty based on combined evidences have been published in [15-16]. Whereas the models in these studies strictly focus on diagnostic procedures, our modeling approach dwells on a monitoring tool that can automatically generate informative clinical data to assist health professionals in the screening of suspected respiratory conditions. We consider this as a major contribution from the study.

In what follows, we throw more light on the rationale for adopting certainty model for the DSS inferencing on the contextual data, as elucidated in the next section. A brief description of the monitoring system design is provided in section 3. Section 4 focuses on the approach for the certainty model while we describe an application scenario of the model in section 5. The last section draws a summary on the discussion and further work.

## 2. Background

### 2.1. Context Recognition

Context recognition in mHealth setting is an active research area that have generated varying interests given the diversity in the components and procedures of capturing contextual data. Multi-modal sensing from embedded sensors in smartphones provides a good platform for high quality context recognition [17].

In the context of exercised-induced respiratory conditions, the principal event being monitored is the audio symptoms extracted from the recordings by the built-in microphone; and the immediate context provided by the mobile system are the activity level from the embedded accelerometer sensor and ambient conditions from temperature and humidity sensors. The mappings of the contextual information polled from various sensors to real-time situations, are often performed using information and probability theories [14, 17]. However, in scenarios where the required information or data to make an inference is not fully available, then alternative techniques are considered. In such cases, the certainty theory becomes handy as alternative approach for inexact reasoning based on available evidences [18].

Certainty theory originates from complexities involved in the development of MYCIN in dealing with inexact information and inexact inference [18]. It is considered a well-known alternative to probability theory for handling inexact reasoning in practical situations as it relates to many expert system applications. It focuses on the heuristics (what) of a task to be accomplished rather than the procedures (how) to accomplish it. In other words, it relies on the judgmental measures of belief on the available evidence and sets aside the more rigorous constraints of probability estimates [18].

### 2.2. Certainty Theory vs. Probability Theory

Probability theory has been described as an ideal technique for dealing with inexact information. And it is most suitable in such areas as weather forecasting and financial planning [18]. However, its consideration as an effective technique in expert systems requires meeting certain constraints which include:

(i) Prior probabilities must be known – requires availability and reliability of statistical information (lack of past or background data limits the use of classical probability approach in medicine).
(ii) Total probability must equal unity – strict probability of Bayesian theorem, i.e. $P(H/E) + P(\sim H/E) = 1$, where H is the hypothesis or proposition and E is the evidence. This is impracticable where there is no relationship between E and $\sim H$.
(iii) Probabilities must be updated - requires recalculating all probabilities in the case of new evidence.
(iv) Conditional Independence is required – This is also not realistic in medical or clinical situations where evidences or symptoms are not mutually exclusive. (E.g. coughing/wheezing due to breathing cold/dry air because of low temperature/humidity; while/after exercising vigorously).

Certainty theory on the other hand, though stems from the probability theory, does not strictly follow the formalities of probability concepts. Comparing both theories, we can establish the following relationship:

The certainty theory in its construct, "suggests that the prior probability of hypothesis H given as P(H) represents expert's critical view of H". Thus, the expert's disbelief in H given as P(~H) is then surmised in line with the conventional probability constraint that the probability 'for' and 'against' a hypothesis sums up to 1 (i.e. $P(H/E) + P(\sim H/E) = 1$) [18].

The theory further postulates that "if the expert observes an evidence E, such that the probability given the evidence (conditional probability) – P(H/E), is greater than the prior probability P(H), then the expert's belief in the hypothesis increases" by the factor: P(H/E) – P(H)/1 - P(H). "On the other hand, if E produces a probability in the hypothesis that is less than P(H), then the expert's belief in the hypothesis will decrease" by: P(H) – P(H/E)/P(H) [18].

Summarily, certainty theory asserts "that a given piece of evidence can either increase or decrease an expert's belief or disbelief in a hypothesis"; and this variability can be expressed as a "measure of belief (MB)" or "measure of disbelief (MD)" [18].

Let $h$ denotes P(H), $he$ denotes P(H/E), $b$ denotes MB, and $d$ denote MD; $0 \leq b \leq 1$, and $0 \leq d \leq 1$

Thus, $b$ and $d$ can be expressed as:

$$b = \begin{cases} 1, & if\ h = 1 \\ \dfrac{\max[he, h] - h}{1 - h}, & otherwise \end{cases} \quad (1)$$

$$d = \begin{cases} 1, & if\ h = 0 \\ \dfrac{\min[he, h] - h}{-h}, & otherwise \end{cases} \quad (2)$$

Certainty theory also accommodates scenarios where contextual data relating to the hypothesis may be observed. In this case, the theory considers the effect of multiple evidences on the belief and disbelief in the proposition; and hence, introduces a new variable known as certainty factor (CF) – 'a net number' that reflects the overall belief in the proposition [18]. Thus, CF is given as:

CF = $b$ - $d$.

The CF value -1 connotes a 'definitely false' hypothesis, 0 means 'unknown hypothesis, while the CF value 1, is interpreted as a 'definitely true' hypothesis [17] as illustrated in Figure1.

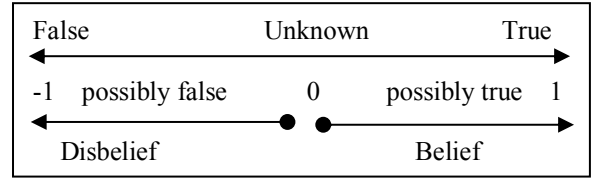

**Figure 1. Range of certainty factor value [18]**

Though the CF equation originally derives from probability estimates, its concept however, does not conform to the analysis and rigidity of probability measure. For instance, the CF values for and against a hypothesis with a given evidence, do not sum to unity compared to the probability for and against a probability that must be equal to 1; i.e. CF(he) + CF(~he) ≠ 1. There also exists conditional certainty factor but its evaluation contradicts the principle of normal conditional probability in cases where evidences are not independent.

From the on-going arguments, we can conclude that while certainty theory may have some elements of probability in its development, the equations defining the associated variables namely – MB, MD, and CF are unstructured (ad hoc), and do not strictly adhere to the formalities of probability theory. Thus, these variables particularly the CFs, are not to be treated as probabilities given that the idea behind the concept is to model or factor in the dynamics of human reasoning, when presented with different scenarios of limited evidences. In other words, CF models are "designed to mimic inexact reasoning of humans" [18].

## 3. The Monitoring System Design

Figure 2 shows the functional components of the proposed monitoring system with emphasis on the certainty model of the inference mechanism. A detailed description of the system design is provided in [19].

The built-in sensors in android phone (Galaxy S4) – microphone, thermometer, hygrometer and accelerometer, automatically collect raw physiological, ambient, and activity data which are then processed and transformed into clinical data using standard signal processing and machine learning techniques [7, 19]. The most common indicators of exercise-induced respiratory conditions as identified by experts are breath sound symptoms – cough, wheeze, stridor, etc. (see Figure 3a).

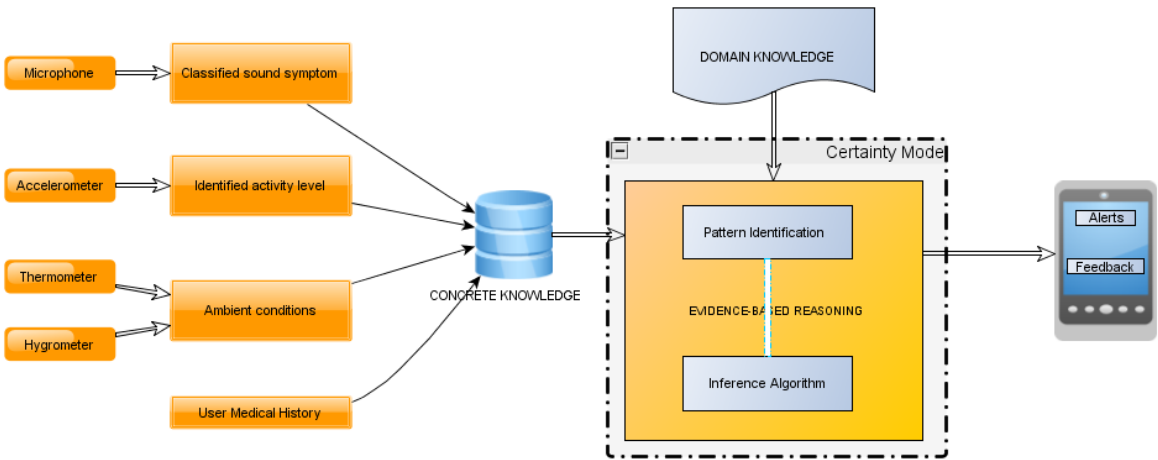

**Figure 2. A schema of the Mobile Monitoring System**

However, these symptoms are not enough evidence to conclude on a suspected respiratory ailment. There is the need to factor in other variables namely the triggers – physical activity level (Figure 3b) and the ambient conditions – temperature and humidity. Whereas sound symptoms are the principal events being monitored, the triggers provide contextual information on detected abnormal respiratory sound which adds to the degree of belief in the suspected illness. The principal events and the contextual evidences constitute the Concrete Knowledge for the reasoner. This knowledge is captured in an embedded database (SQLite) as shown in Figure 3c.

The medical knowledge used in modeling the reasoner were elicited from interaction with experts in pulmonology and sports medicine. Other sources of knowledge include literature and interviews with persons affected by the respiratory conditions under consideration. These altogether form the Domain or Abstract Knowledge fed into the certainty model. The model provides the monitoring system the intelligence to reason on the evidences based on the domain knowledge; in order to alert the user or generate feedback on the user's health status. The generated information or report can be stored on the mobile device or shared with healthcare providers and physicians for subsequent actions.

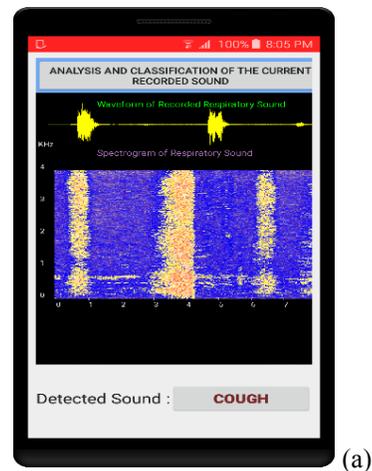
(a)

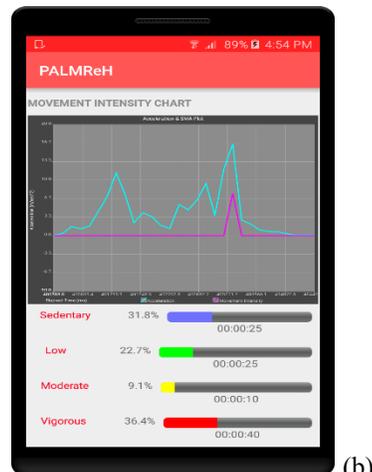
(b)

(c)

**Figures 3 (a) – (c). Concrete knowledge automatically captured by the mobile monitoring system**

## 4. The Certainty Model

The goal of the automated mobile decision support system is to alert patients and their caregivers on detection of abnormality in the measured variables that could predispose respiratory distress. But the overlap in the symptoms and triggers of the respiratory conditions complicates the reasoning process in discriminating the evidence for each of the suspected conditions; and thus, would require obtaining more detailed knowledge from experts to confirm or exclude a suspected respiratory condition.

In view of the above constraints, we adopt certain theory to model the inference mechanism of the monitoring system based on the available evidences. Our approach deviates from the conventional rule-based models that assume a 'single-fault' (cause) i.e. only one disorder is suspected from the given evidences. We rather consider a 'multi-fault' assumption where more than one disorder or respiratory conditions is suspected based on the analysis of observed variables, and the certainty weight of the suspected conditions [15,16]. The concepts of certainty theory are applied to both knowledge representation and the reasoning process.

### 4.1. Knowledge Representation

To formalize the knowledge extracted from the problem domain, we consider mathematical expressions in the form of Set notations to represent the respiratory conditions, their symptoms and triggers in the problem space, as well as the participation ratio of the aforementioned variables. Other forms of knowledge representation exist as described in [20,21]. However, we adopt an event-driven approach [21] for the development of the certainty model since we are considering an automated decision support system. First, we start with the definitions of measurable or observable factors within a range of variation.

Let C denotes a set of respiratory conditions a patient may have:

$$C = \{c_1, c_2, \ldots, c_n\} \qquad (3)$$

Let S(c) represents a set of possible symptoms of condition c:

$$S(c) = \{s : s \text{ is a symptom of } c, c \in C\} \qquad (4)$$

Let T(c) also denotes a set of known triggers of the respiratory conditions:

$$T(c) = \{t : t \text{ is a trigger of } c, c \in C\} \qquad (5)$$

Let P be a universal set of all observable symptoms and context (triggers) signaling a respiratory condition:

$$P = \bigcup_{c \in C} (S(c), T(c)) \qquad (6)$$

Let Q be a set of symptoms and triggers detected or observed by the monitoring system:

$$Q = \{q : q \text{ is an observed evidence}, q \in P\} \qquad (7)$$

β represents a set of suspected conditions characterized by the detected events or observed evidences:

$$\beta = \{c : c \in C, (Q \cap P) \neq \emptyset\} \qquad (8)$$

The certainty of a suspected condition is then determined by the presence of its symptoms and triggers. It is calculated by the participation ratio of each of the symptoms and triggers - $R_s$ and $R_t$ respectively.
Let $R_s(c)$ be the participation rate of symptom $s$ in condition $c$, and $R_t(c)$ be the participation rate of trigger $t$ in condition $c$, then:
γ(c) = Max {$R_s(c)$ : s ∈ S} and
θ(c) = Max {$R_t(c)$ : t ∈ T}

Thus, the certainty weight $W(c)$, for each of the suspected condition is computed based on the participation ratios of the combined evidences and the asymptotic property of 'Incremental Acquired Evidence' in the certainty theory/model [18].

$$W(c) = \frac{\gamma(c) + \theta(c)}{\max\,[(\gamma(c) + \theta(c)), (1 - \gamma(c) * \theta(c))]} \qquad (9)$$

Then Φ being the condition with the largest certainty weight $W(c)$ is selected from the list of suspected conditions β.

The parameters γ and θ are determined based on the disjunctive rule of "Certainty Propagation for Multiple Premise Rules" in the Certainty Theory construct [18]. Equation 9 stems from "Certainty Propagation for Similarly Concluded Rules" of the same theory. However, the normalization factor $\max[(\gamma(c) + \theta(c)), (1 - \gamma(c) * \theta(c))]$ in our model was introduced for two reasons: First, in cases where the evidences are not totally independent or mutually exclusive as noted in section 2.2, which we addressed here by using the expression $1 - \gamma(c) * \theta(c)$. Second, to mitigate the effect of $\gamma(c) * \theta(c) \geq 1$. The essence is to maintain positive values for the certainty weights as highlighted in the CF scale (Figure 1), and to satisfy the asymptotic property of the certainty model; since the proposed monitoring system is primarily concerned with positive evidences to support the belief in the suspected conditions.

### 4.2. The Reasoning Process

From the knowledge representations expressed above, we can imbue the monitoring system with some level of intelligence through ad hoc imprecise reasoning, using the following light-weight inference algorithm as described thus:

First, the system accepts as input all the static variables compiled from the domain knowledge and stored as xml file. The model then initializes all the required variables. All respiratory conditions $C$ in the domain space are initially suspected. Likewise, all symptoms and triggers $P$ in the domain space are initially expected. $Q$ is also initialized to an empty set and then populated by reading periodically (daily), observed evidences from the SQLite database as captured by the monitoring system. The system then determines β based on $Q$ by checking if $Q$ intersects with P. If so, the system prunes β by removing respiratory conditions whose evidences were not captured in $Q$. $R_s$ and $R_t$ of each element in $(Q \cap P)$ is calculated. The parameters γ and θ (representing the maximum of $R_s$ and $R_t$ for each suspected condition in β) are then determined. Using these two parameters, the system computes the certainty weight W(c) of each suspected condition by applying equation 9, and finds Φ being the suspected condition(s) with the largest certainty weight; It then displays Q, Φ, and W(c) on the monitoring device's screen and also provides feedback and/or alert to the user that can be shared with care givers and concerned health professionals. Figure 4 illustrates the flow diagram of the reasoning process.

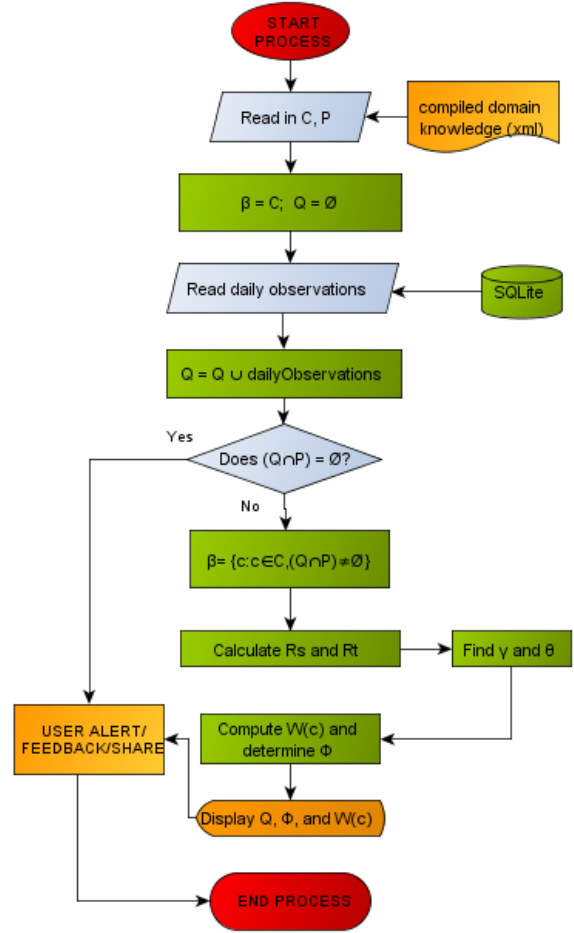

**Figures 4. Flow Diagram of the Reasoning Process**

## 5. Domain Scenario

### 5.1. Clinical Application

Here, we describe an application case of the certainty model. Consider a scenario where a patient has earlier been diagnosed with a certain respiratory condition– with symptoms and triggers well documented and made known to him. Assume the patient is a high school student that loves athletics and as such, trains with his peers from time to time. Based on his condition, he has been advised on the type of physical activity to engage in, and certain weather conditions to avoid. However, due to peer pressure or concerns about inadequacy or incompetence, he does not adhere to instructions by his physicians, and continues to engage in high-risk exercises under adverse ambient conditions that could flare up or trigger symptoms of the respiratory condition. Suppose he wears the monitoring device on him, and the system is furnished with the following domain and concrete knowledge as defined in the equations and the algorithms in section 4. Assume the evidences observed by the monitoring device were wheezes, cough, low temperature, low humidity, and vigorous exercise abbreviated as whz, cgh, lt, lh, and vgr respectively; then the set Q is populated as:

$$Q = \{whz, cgh, lt, lh, vgr\}$$

From the domain knowledge, the following set of respiratory conditions which are associated with the observed evidences (triggers and symptoms) can be identified as shown in Table 1.

**Table 1. Suspected Conditions based on observed evidences**

| Resp. Conditions | Evidences |
|---|---|
| EIA | Wheeze, Cough, Low temperature, Low humidity, Vigorous exercise |
| EIB | Wheeze, Cough, Low temperature, Low humidity, Vigorous exercise |
| VCD | Cough, Low temperature, Low humidity, Vigorous exercise |
| EIR | Low temperature, Low humidity, Vigorous exercise |
| COPD | Cough, Low temperature |

With the information in Table 1 we can calculate the participation ratios of the observed evidences (w.r.t number of suspected conditions associated with each symptom and trigger) as follows:

**Table 2. Participation Ratios of Observed Evidences**

| Evidences (Q) | Participation Ratios |
|---|---|
| Wheeze | 0.5 |
| Cough | 0.25 |
| Low temperature | 0.2 |
| Low humidity | 0.25 |
| Vigorous exercise | 0.25 |

Thus, using Equation 9, the certainty weights for the suspected respiratory conditions are computed as:

| | | |
|---|---|---|
| EIB | = | 0.85 |
| EIA | = | 0.85 |
| VCD | = | 0.53 |
| EIR | = | 0.25 |
| COPD | = | 0.47 |

The implications of the results obtained from the computations can be viewed or interpreted based on the case being analyzed and the evidences captured by the monitoring device. First we observe that EIR has the least suspicion index of 0.25, though its triggers were among the evidences, the sound symptoms (sneeze and sniffle) were not captured; and as such, it is automatically excluded from the list of suspected conditions. COPD is not triggered by vigorous exercise but the sound symptoms were among the evidences and the exacerbation is characterized by low physical activity due to low temperature [22-23]. However, it also qualifies as a candidate for exclusion based on its certainty weight which is 0.47. VCD ranks second in the list with certainty weight of 0.53, the reason being the resemblance of its triggers and symptoms to those of asthma [24]. The imprecision in the evidences however, calls for further probe. EIB and EIA have the same and highest suspicion index – 0.85, which intuitively is no surprise, given that both terms are often used interchangeably in the problem domain; as there are no differences in the symptoms manifested by the conditions when provoked by physical activity. However, to create a distinction in the terminologies, health experts define EIA as respiratory distress in persons diagnosed with asthma condition predisposed by physical exertion; while EIB refers to the manifestation of asthma-like symptoms in non-asthmatics only during rigorous exercise like sporting activities [6, 25].

In addition to the measurable evidences, the system may adopt a status of medical history/diagnosis of any of the conditions for a patient, which is provided by the user at the profile setup of the monitoring application. This information being part of the concrete knowledge as shown in Fig. 2, can be used for discrimination between EIA and EIB. Also, based on the knowledge extracted from domain experts, the timing of the occurrences of the symptoms with respect to the duration of vigorous exercise is considered important in the differential diagnosis of EIB and VCD. EIB wheezes are known to occur after the first 5 minutes of vigorous exercise and may persist up to 10 minutes after physical exertion [6]. VCD stridor on the other hand, is observed with intense or vigorous exercise [24]. This information can be deduced from the event time as captured in the database table in Fig. 3c. Inclusion of the additional knowledge in the certainty model (implemented by rules), will consequently raise or lower the suspicion indices of the respiratory conditions.

We would like to point out here that both symptoms and triggers are not so sufficient for a comprehensive respiratory diagnosis in clinical settings. Lung function test is considered important in the differential diagnosis of the respiratory conditions [6]. Dyspnea or shortness of breath is another critical sign of respiratory distress which can be modeled from the respiratory rate of a patient. We hope to include these two measurements as special modules in the monitoring system as our study progresses. However, the usefulness of the proposed system is in generating objective data that can provide decision support to health professionals in the pre-clinical screening and diagnosis of a patient during office visits.

## 5.2. Clinical Evaluation

We intend to carry out a pilot study that will involve both the experts and patients in the study domain. The medical experts will help provide assessment of patients and their eligibility to participate in the study.

At first, we will consider asymptomatic patients (individuals who have been diagnosed with respiratory conditions but do not show symptoms or signs of respiratory distress at the time of conducting the study); where smartphones will be used to capture their breath sounds while they voluntarily and with the approval of a physician, engage in different types of activities (climbing stair case, jogging, walking, sitting, etc.). The study will require subjects wearing the monitoring device while performing short burst of activities within a specific period (2 to 5 minutes). The system takes measurements of individual's breath sound and activity patterns concurrently at specified time intervals. The recording may be performed both indoors and outdoors in order to measure environmental data such as temperature and relative humidity. The study population will be exclusively adults within the age range of 20-48 years with a sample size of about 10-20.

Since the main purpose of the field study is to validate the functional components of the proposed system, subjects will be asked to fill a form to indicate their experiences with their respiratory conditions, and the health personnel in charge will be asked to confirm the status of the participant's respiratory conditions to help correlate and corroborate the data being collected by the monitoring system.

## 5.3. Patient's Self-management

We have partially implemented an aspect of proposed monitoring system that will enable patients keep track of the symptoms and triggers of their respiratory conditions. This module of the system shows the user a visualized summary of the captured events on daily basis using column charts for measuring the frequency of symptom occurrence at specific periods of the day, a line chart to measure variations in the ambient temperature and relative humidity, and the bubble chart to display the intensity aggregate of each activity level at given hours of the day. This essentially provides the user with basic information to correlate the measured events. For example, knowing the specific period of the day when a symptom gets worse and the triggers that aggravate the symptoms, would help the patient to personally manage and control his/her condition.

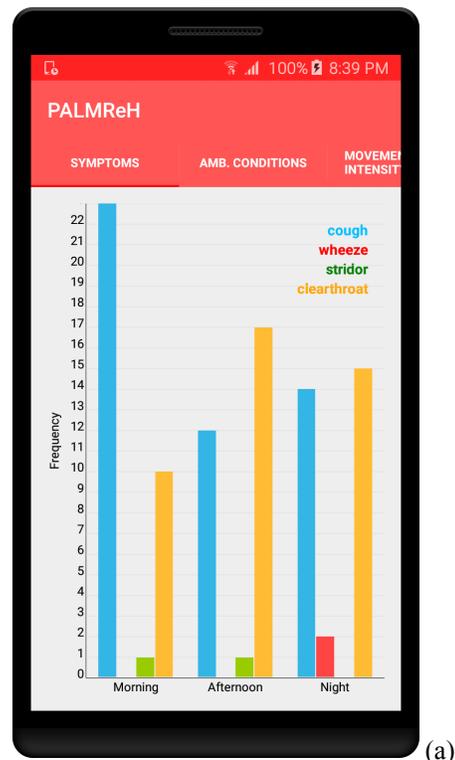

(a)

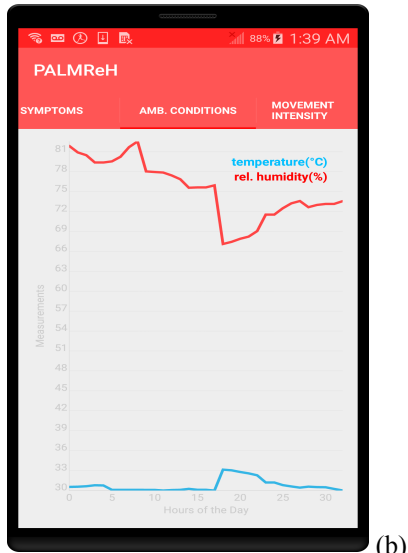
(b)

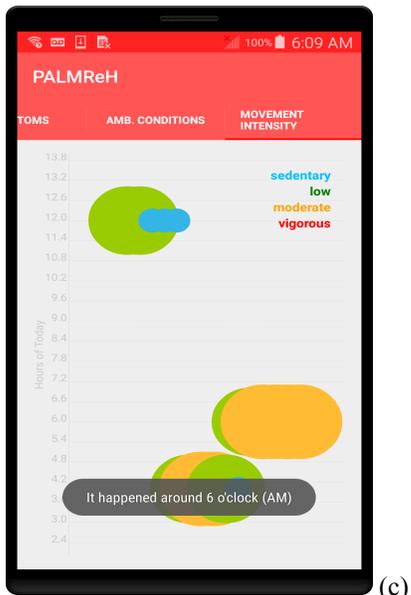
(c)

**Figures 5 (a-c). Daily events summary by charts**

Figures 5 (a - c) demonstrate preliminary tests of the daily events summary. The monitoring system is designed and developed to run as a real-time standalone system on an android device with no recourse to any backend system. So far, we have had an appreciable response time and low power usage in the execution of both individual and integrated modules (sensors' data capturing and processing, machine learning for detection of sound symptoms, storing and retrieving data from the embedded SQLite database) of the monitoring system. We hope to provide more details on device resource usage when we fully implement the system.

## 6. Conclusions

The proposed certainty model for the monitoring system is patterned after Clinical Decision Support Systems. However, it is not necessarily considered a diagnostic tool but an assistive tool. While its major goal is to alert users on detection of any anomaly and to provide immediate feedback on their health status; it is also aimed at assisting health professionals with objective data in patient's screening of a suspected respiratory ailment. The multi-level observations from various components of the system form the basis for a follow–up on the reported patient's status. The model implements a light-weight inference algorithm intended for ad hoc mobile monitoring. We adopted this approach in our modeling since the 'frame of discernment' in the study domain does not preclude comorbidity cases among suspected respiratory conditions. Thus, evidences captured can be pointers to more than one ailment which can be confirmed or refuted by conducting further clinical tests and observations on the patient. We are aware of other decision-making approaches for modeling inexact reasoning such as Dempster–Shafer theory (DST) and the transferable belief model (TBM). In our future study, we will evaluate the proposed model alongside these other alternative frameworks of reasoning, and compare their performances in a decision-making setting.